\documentclass{article}
\usepackage[preprint]{colm2026_conference}

\usepackage{microtype}
\usepackage{hyperref}
\usepackage{url}
\usepackage{booktabs}
\usepackage{graphicx}
\usepackage{colortbl}
\definecolor{oursblue}{RGB}{219,234,254}

\definecolor{darkblue}{rgb}{0, 0, 0.5}
\hypersetup{colorlinks=true, citecolor=darkblue, linkcolor=darkblue, urlcolor=darkblue}

\title{GRP: Goal-Reversed Prompting for Zero-Shot\\
Evaluation with LLMs}

\author{Mingyang Song, Mao Zheng, Xuan Luo\\
Tencent, China\\
\texttt{nickmysong@tencent.com}}

\begin{document}

\maketitle

\begin{abstract}
Pairwise LLM-as-a-judge evaluation asks the judge to identify the
\emph{better} of two candidate answers. We study a one-line
modification that asks for the \emph{worse} answer instead and
recovers the preference by elimination, a procedure we call
Goal-Reversed Prompting (GRP). GRP introduces no extra inference
rounds, composes with any prompt template (direct, chain-of-thought,
or Arena-Hard SOP), and leaves the rest of the evaluation pipeline
untouched. Two observations motivate the reversal. Reverse reasoning
is a recurring strategy in human problem solving, and modern
instruction-tuned judges exhibit a positive-leaning bias that asking
for the worse answer can counteract. On JudgeBench under a strict
consistency protocol that counts a judgment as correct only when both
response orderings agree with the gold preference, GRP improves all
three closed-source judges we test across both response-pair sources.
With GPT-4o-generated pairs, the Arena-Hard SOP baseline improves
from 61.71\% to 66.23\% for GPT-4o (+4.52) and from 60.00\% to
66.00\% for Claude-3.5-Sonnet (+6.00), with the largest absolute
gains on Reasoning and Mathematics. The lift persists when response
pairs come from Claude-3.5-Sonnet and when the SOP scaffolding is
stripped to a minimal direct-prompting template, suggesting that goal
reversal acts on the underlying judging behavior rather than on a
particular rubric. Stronger judges benefit more than weaker ones,
suggesting that goal reversal exposes additional
reasoning capacity rather than compensating for its absence.
\end{abstract}

\section{Introduction}

LLM-as-a-judge evaluation has become a common substitute for human
preference judgments because collecting annotations at the scale that
modern training pipelines require is slow and expensive
\citep{judge,judgebench}. The dominant instantiation is
\emph{pairwise} judgment, where a judge model reads a user prompt
together with two candidate answers and returns a preference
\citep{judge,unfair}. A large body of work has examined failure
modes of such judges, including positional bias, self-preference, and
verbosity bias \citep{unfair,chen,many-shot-judge}, and has proposed
corrections that act on the judging procedure, on the judge's
training data, or on post-hoc aggregation.

This work starts from a different angle. Rather than modifying
the judge or the aggregation, we revisit a small and usually
unexamined choice inside the prompt itself, the direction in which the
judge is asked to state its preference. The standard phrasing asks
the judge to identify the \emph{better} answer. We ask whether
requesting the \emph{worse} answer and recovering the preference by
elimination behaves differently under an otherwise identical protocol.
We call this procedure \textbf{Goal-Reversed Prompting} (GRP).

Two independent observations motivate the reversal.
Reverse reasoning is a recurring strategy in human problem solving,
particularly in mathematics and decision making where the weaker
option has a more salient flaw than the stronger one has a
distinguishing virtue \citep{reverse_thinking,reverse_think}.
Separately, modern instruction-tuned LLMs tend toward
positive-leaning, agreeable judgments, and prior work has documented
systematic bias toward one side of a pair \citep{unfair}. Asking for
the worse answer tests whether a fraction of pairwise-judge errors
comes from the difficulty of \emph{endorsing} a winner rather than
from the underlying ability to distinguish the two candidates. We do
not claim to disentangle these two forces here, and treat them as
hypotheses that guide our design rather than as mechanisms to be
proved.

To make the comparison fair, we keep everything except the goal
sentence identical between conditions. The judge model, prompt
scaffolding, input candidates, decoding configuration, and
order-swap procedure are all shared. The only textual change is
replacing \emph{better} with \emph{worse} in the goal sentence and
inverting the two verdict labels. We adopt a strict consistency
protocol in which each instance is evaluated twice with the two
candidate answers swapped, and a judgment counts as correct only if
both trials agree with the gold preference after un-swapping. Ties
and order-dependent flips count as incorrect. Because this protocol
applies uniformly to both conditions, the comparison is
order-of-presentation invariant by construction.

We report results on JudgeBench \citep{judgebench}, using both its
GPT-4o-generated and Claude-3.5-Sonnet-generated response-pair
versions, and three closed-source judges (GPT-4o, Gemini-1.5-Pro,
Claude-3.5-Sonnet). Under the Arena-Hard SOP template on
GPT-4o-generated pairs, GRP lifts overall accuracy from 61.71\% to
66.23\% for GPT-4o, from 57.14\% to 59.43\% for Gemini-1.5-Pro, and
from 60.00\% to 66.00\% for Claude-3.5-Sonnet. The gains carry over
to the Claude-3.5-Sonnet response-pair version and to a stripped-down
direct-prompting template, where every judge gains \textbf{4--5}
points overall despite the absence of the SOP rubric. The absolute
gains concentrate on Reasoning and Mathematics, where the gold
preference is closer to a verifiable correctness signal. The stronger
judges (GPT-4o 2024-11-20 and Claude-3.5-Sonnet) benefit more than
the weaker one, a pattern suggesting that reversal leverages
reasoning capacity the judge already possesses. A judge that
cannot reliably verify candidate correctness in the forward direction
benefits little from posing the question in reverse.

\section{Goal-Reversed Prompting}
\label{sec:method}

\begin{figure}[t]
\centering
\includegraphics[width=0.60\linewidth]{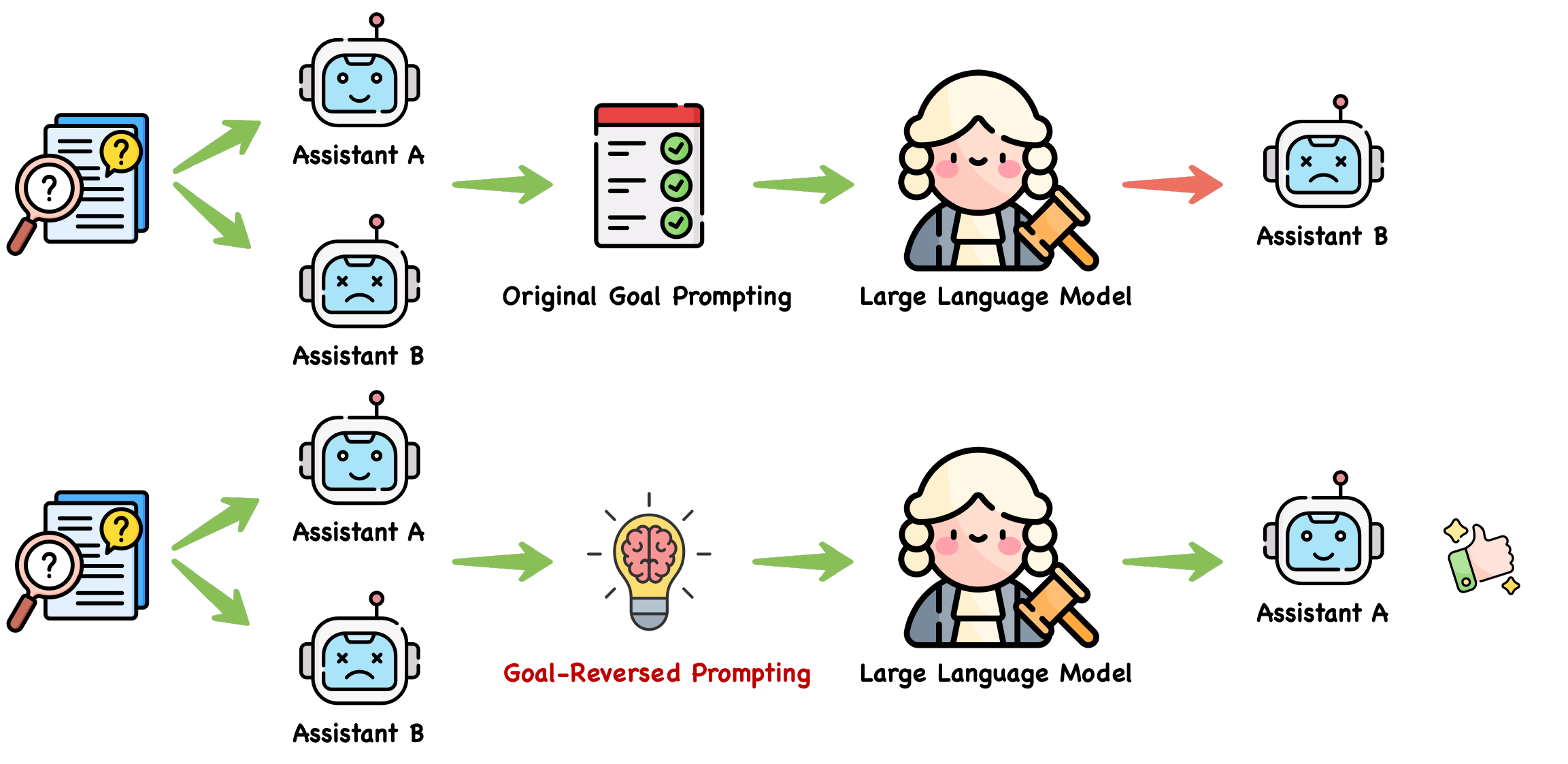}
\caption{Pairwise comparison under the original goal (top) and under
the reverse goal (bottom). The two procedures differ only in the
final question posed to the judge and the label that encodes the
verdict.}
\label{fig:intro}
\end{figure}

\subsection{Setup}

A pairwise judgment instance consists of a user prompt $x$ and two
candidate answers $a$ and $b$. A judge model $J$, conditioned on a
template $T$, outputs a verdict in $\{a{>}b, b{>}a, a{=}b\}$. The
standard \emph{original-goal} template $T_{\mathrm{orig}}$ instructs
$J$ to identify the \emph{better} answer. We define the
\emph{reverse-goal} template $T_{\mathrm{rev}}$ as $T_{\mathrm{orig}}$
with two local changes: (i)~the goal sentence asks which answer is
\emph{worse}, and (ii)~the verdict labels are inverted, so that
``Assistant~A is worse: [[B$>$A]]'' replaces ``Assistant~A is better:
[[A$>$B]]''. Everything else remains identical.

To reduce positional bias \citep{judge,unfair}, each instance is
evaluated twice with $a$ and $b$ swapped. A judgment is
\emph{consistent} if the two trials agree after un-swapping.
Inconsistent verdicts and ties are marked incorrect. This protocol
is stricter than the original JudgeBench reporting and applies
identically to both $T_{\mathrm{orig}}$ and $T_{\mathrm{rev}}$.

\subsection{Three Template Variants}

We instantiate GRP on three template families to test how it interacts
with prompt scaffolding:
\begin{itemize}
    \item \textbf{Direct.} The judge is asked to pick the better
    (resp.\ worse) answer with no chain-of-thought and no scoring
    rubric.
    \item \textbf{CoT.} The judge is additionally instructed to think
    step by step before producing a verdict.
    \item \textbf{SOP (Arena-Hard).} The judge is given a structured
    rubric. It first generates its own reference answer, then compares
    both candidate answers against it, evaluates helpfulness, relevance
    and conciseness, and finally outputs the verdict
    \citep{judgebench}.
\end{itemize}
Table~\ref{tab:zero_shot} shows all three variants under the reverse goal.
The original-goal counterparts follow the same surface edit in reverse
and are listed in Appendix~\ref{app:templates}.

\begin{table*}[t]
\centering
\small
\renewcommand\arraystretch{1.15}
\resizebox{0.99\linewidth}{!}{
\begin{tabular}[t]{@{}p{0.99\linewidth}@{}}
\toprule
\textbf{Direct prompting (reverse goal)} \\
\midrule
Please act as an impartial judge and evaluate the quality of the
responses provided by two AI assistants to the user prompt displayed
below. You will be given assistant A's answer and assistant B's answer.
Your job is to evaluate which assistant's answer is \textbf{worse}, and
output one of the following choices as your final verdict with a label.
1. Assistant A is \textbf{worse}: [[B$>$A]].
2. Assistant B is \textbf{worse}: [[A$>$B]].
Example output: ``My final verdict is Assistant A is \textbf{worse}: [[B$>$A]]''. \\
\midrule
\textbf{CoT prompting (reverse goal)} \\
\midrule
Same as above, with an additional final instruction: ``Let's think step
by step.''. \\
\midrule
\textbf{Arena-Hard SOP prompting (reverse goal)} \\
\midrule
Please act as an impartial judge and evaluate the quality of the
responses provided by two AI assistants. Your job is to identify which
assistant's answer is \textbf{worse}.
Begin by generating your own answer to the prompt before judging the
candidates. When evaluating, compare both candidates against your own
answer, identify and correct any mistakes, then assess helpfulness,
relevance, and conciseness, and finally consider creativity and missing
information.
After your explanation, output one of the following choices as your
final verdict with a label.
1. Assistant A is \textbf{worse}: [[B$>$A]].
2. Assistant B is \textbf{worse}: [[A$>$B]].
Example output: ``My final verdict is Assistant A is \textbf{worse}: [[B$>$A]]''. \\
\midrule
\textbf{Inputs (shared across all three variants)} \\
\midrule
$\langle$User Prompt$\rangle$ \{User-Prompt\} \\
$\langle$The Start of Assistant A's Answer$\rangle$ \{Answer-A\} $\langle$The End of Assistant A's Answer$\rangle$ \\
$\langle$The Start of Assistant B's Answer$\rangle$ \{Answer-B\} $\langle$The End of Assistant B's Answer$\rangle$ \\
\bottomrule
\end{tabular}}
\caption{Three GRP template variants. Original-goal counterparts
are obtained by replacing every \emph{worse} with \emph{better}
and inverting the two verdict labels.}
\label{tab:zero_shot}
\end{table*}

\subsection{Two Hypotheses}

We state two hypotheses about why goal reversal may help, without
attempting to prove either. The experiments in
Section~\ref{sec:results} probe which hypothesis better fits the
observed pattern of gains.

\textbf{H1 (rule-out is easier than rule-in).} Under the original
goal, the judge must commit to one answer as better. When both
candidates are plausible, this requires reasoning about a fine-grained
advantage that may be small relative to surface variation in style.
Under the reverse goal, the instance is resolved once the judge
identifies a single salient flaw in one candidate. Where the
gold preference is closer to a verifiable correctness signal than to
a stylistic preference, ruling one candidate out is a more direct
path to the verdict. The Reasoning and Mathematics subsets of
JudgeBench \citep{judgebench} fit this profile, and the lift we
observe is largest on these subsets.

\textbf{H2 (counteracting positive-leaning bias).} Modern
instruction-tuned LLMs are trained to produce helpful, agreeable
outputs and tend toward positive-leaning judgment behavior in
pairwise comparison \citep{judge,unfair}. The original goal aligns
with that tendency because it asks the judge to endorse a winner. The
reverse goal asks the judge to flag a loser, which does not align
with the positive-leaning default. To the extent that the
original-goal error budget includes over-endorsing weak candidates,
flipping the question reduces the alignment between the bias and the
verdict label at no computational cost. We do not separate H1 and H2
in this paper and leave a clean ablation (e.g., a setting where both
candidates are weak) to future work.

\section{Experimental Setup}
\label{sec:setup}

\paragraph{Benchmark.}
We use JudgeBench \citep{judgebench}, which evaluates a judge's
ability to identify the objectively better of two candidate answers.
We test on two response-pair sources, the GPT-4o version (350
instances across 154 Knowledge, 98 Reasoning, 56 Math, 42 Coding) and
the Claude-3.5-Sonnet version (270 instances across 154 Knowledge, 51
Reasoning, 34 Math, 31 Coding). Following JudgeBench, we report
accuracy per category and an overall score.

\paragraph{Judges.}
Our main results use three closed-source judges accessed through their
official APIs: GPT-4o (versions 2024-05-13, 2024-11-20),
Gemini-1.5-Pro (versions 001, 002), and Claude-3.5-Sonnet
(2024-06-20). We additionally report GPT-4o-mini (2024-07-18),
Gemini-1.5-Flash (001), and Claude-3-Haiku (2024-03-07) under the
original goal re-evaluated with our strict protocol, but omit
reverse-goal runs because their baselines are weak.

\paragraph{Decoding.}
All judges use greedy decoding (temperature $=0$) for reproducibility.
We do not impose a length limit beyond the API default.

\paragraph{Strict consistency protocol.}
As defined in Section~\ref{sec:method}, each instance is evaluated
twice with response order swapped, and only judgments that agree on
both trials count as correct. This protocol is stricter than that of
\citet{judgebench}, which is one reason some original-goal numbers we
report differ from theirs (marked $^\dagger$).

\section{Results and Analysis}
\label{sec:results}

\subsection{Main Results}

\begin{table*}[t]
\centering
\small
\renewcommand\tabcolsep{8pt}
\renewcommand\arraystretch{1.20}
\begin{tabular}{lcccccc}
\toprule
\textsc{LLMs} & \textit{Version} & \textit{Knowledge} & \textit{Reasoning} & \textit{Math} & \textit{Coding} & \textbf{\textit{Overall}} \\
\midrule
\multicolumn{7}{l}{\textit{SOP prompting (original goal)}} \\
\midrule
\textsc{GPT-4o-mini}$^\dagger$       & \scriptsize\textit{2024-07-18} & 46.75 & 42.86 & 66.07 & 42.86 & 48.29 \\
\textsc{GPT-4o}$^\dagger$            & \scriptsize\textit{2024-05-13} & 46.10 & 48.98 & 62.50 & 52.38 & 50.29 \\
\textsc{GPT-4o}                      & \scriptsize\textit{2024-11-20} & 61.04 & 56.12 & 73.21 & 61.90 & 61.71 \\
\textsc{Gemini-1.5-Flash}$^\dagger$  & \scriptsize\textit{001}        & 26.62 & 20.41 & 30.36 & 9.52  & 23.43 \\
\textsc{Gemini-1.5-Pro}$^\dagger$    & \scriptsize\textit{001}        & 45.45 & 37.76 & 55.36 & 9.52  & 40.57 \\
\textsc{Gemini-1.5-Pro}              & \scriptsize\textit{002}        & 57.79 & 52.04 & 71.43 & 47.62 & 57.14 \\
\textsc{Claude-3-Haiku}$^\dagger$    & \scriptsize\textit{2024-03-07} & 20.78 & 14.29 & 16.07 & 2.38  & 16.00 \\
\textsc{Claude-3.5-Sonnet}$^\dagger$ & \scriptsize\textit{2024-06-20} & 59.74 & 62.24 & 60.71 & 54.76 & 60.00 \\
\midrule
\multicolumn{7}{l}{\textit{SOP prompting (reverse goal, ours)}} \\
\midrule
\rowcolor{oursblue}
\textsc{GPT-4o}             & \scriptsize\textit{2024-11-20} & \textbf{68.18} & 61.22 & 71.43 & \textbf{63.81} & \textbf{66.23} \\
\rowcolor{oursblue}
\textsc{Gemini-1.5-Pro}     & \scriptsize\textit{002}        & 56.49 & 59.18 & \textbf{76.79} & 47.62 & 59.43 \\
\rowcolor{oursblue}
\textsc{Claude-3.5-Sonnet}  & \scriptsize\textit{2024-06-20} & 62.99 & \textbf{66.32} & \textbf{76.79} & 61.90 & 66.00 \\
\bottomrule
\end{tabular}
\caption{JudgeBench accuracy with response pairs generated by
\textbf{GPT-4o}. $^\dagger$ rows are re-evaluated under our strict
two-trial protocol. Bold marks the best per column among the three
reverse-goal judges.}
\label{tab:4o}
\end{table*}

Table~\ref{tab:4o} reports results on the GPT-4o response pairs.
All three judges improve overall when the goal is reversed:
GPT-4o~(2024-11-20) moves from 61.71\% to 66.23\% (+4.52),
Gemini-1.5-Pro~(002) from 57.14\% to 59.43\% (+2.29), and
Claude-3.5-Sonnet~(2024-06-20) from 60.00\% to 66.00\% (+6.00).
The largest absolute gains appear on Mathematics and Reasoning, where
the gold preference is closer to a verifiable correctness signal.
Gemini-1.5-Pro and Claude-3.5-Sonnet both reach 76.79 on Math under
the reverse goal (+5.36 and +16.08 over their respective baselines),
and Reasoning improves from 62.24 to 66.32 (+4.08) for
Claude-3.5-Sonnet. Coding gains are modest for GPT-4o
(+1.91) and roughly flat for the other two judges, whose SOP
baselines on that subset are already low.

A second pattern concerns judge strength. GPT-4o and Claude-3.5-Sonnet,
the two strongest judges under the original goal, also gain most from
the reversal (+4.52 and +6.00), while Gemini-1.5-Pro starts lower and
gains less (+2.29). This gradient supports the hypothesis that
reversal amplifies existing reasoning ability rather than
compensating for its absence. A judge that cannot reliably verify
candidate correctness in the forward direction gains little from
asking the question backwards.

\subsection{Robustness to the Response-Pair Source}

\begin{figure}[t]
\centering
\includegraphics[width=0.60\linewidth]{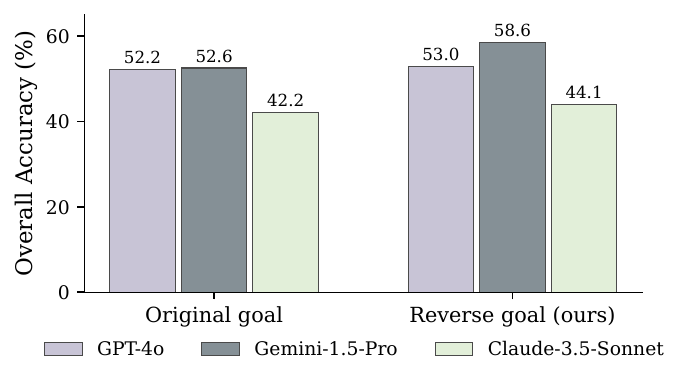}
\caption{JudgeBench overall accuracy with response pairs generated by
\textbf{Claude-3.5-Sonnet}. The reverse goal shifts every judge
upward. Claude-3.5-Sonnet, also the response-pair generator here,
has the lowest accuracy, consistent with reported self-preference
bias \citep{judge,unfair}.}
\label{fig:mainc35}
\end{figure}

When response pairs are generated by Claude-3.5-Sonnet instead of
GPT-4o, the reverse goal still yields a positive shift for every judge
(+0.74 for GPT-4o, +6.00 for Gemini-1.5-Pro, +1.85 for
Claude-3.5-Sonnet, as shown in Figure~\ref{fig:mainc35}).
Claude-3.5-Sonnet has the lowest absolute accuracy on this version,
where it is also the response generator, a pattern consistent
with prior reports of self-preference bias \citep{judge,unfair}.
Its reverse-goal lift is the smallest of the three, though still
positive, suggesting that goal reversal does not remove the
self-preference cue but rather competes with it.

\subsection{Effect of the Prompt Scaffolding}

\begin{figure}[t!]
\centering
\includegraphics[width=0.60\linewidth]{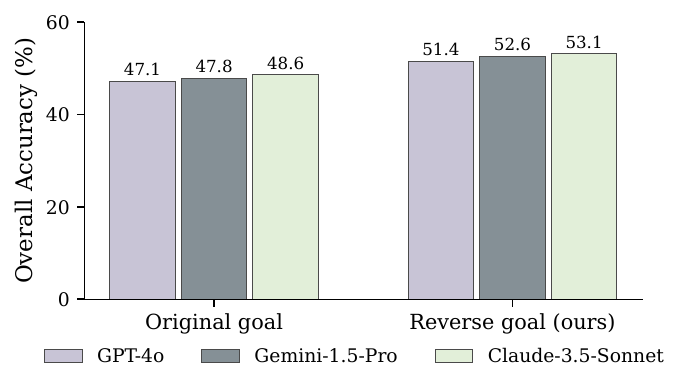}
\caption{Direct prompting on JudgeBench with \textbf{GPT-4o}-generated
response pairs. Removing the SOP scaffolding lowers absolute accuracy
for every judge, yet the reverse-goal lift remains positive across all
three (+4.29, +4.79, +4.57).}
\label{fig:ablation}
\end{figure}

A natural question is whether the gain depends on the Arena-Hard SOP
rubric. Stripping the scaffolding to a minimal direct-prompting
template lowers absolute accuracy for all three judges, as expected,
but the reverse-goal lift persists at +4 to +5 points per judge
(Figure~\ref{fig:ablation}). Together with the robustness results
above, this indicates that goal reversal acts on the underlying
judging behavior and is largely orthogonal to template engineering.

\subsection{Where the Gains Come From}

Across Table~\ref{tab:4o} and Figures~\ref{fig:mainc35}--\ref{fig:ablation}, gains concentrate on Mathematics and Reasoning rather than Knowledge or Coding.
We lack a controlled experiment that isolates the cause, but two observations suggest partial explanations.
First, on Math and Reasoning the gold label reflects objectively verifiable correctness.
Identifying a concrete error in the worse answer is a more direct path to the verdict than arguing why the better answer is preferable, and goal reversal foregrounds exactly this error-finding mode.
Second, the strict consistency protocol penalizes order-dependent judgments, and reversing the goal appears to reduce the fraction of inconsistent verdicts (consistency rates in Appendix~\ref{app:consistency}).
The two effects are confounded in our current design, and disentangling them remains future work.

\section{Related Work}
\label{sec:related}

\paragraph{LLM-as-a-judge.}
LLM-based pairwise evaluation is widely used for ranking model
outputs \citep{judge,judgebench,Koo23,Liusie2023,liu2023,Zhu2023,
lu2023,Fu2023}. Follow-up work has documented failure modes such as
positional bias, self-preference, and verbosity bias
\citep{unfair,chen,many-shot-judge}. GRP does not propose a new
judge architecture or a new debiasing mechanism, and composes with
existing prompt templates unchanged. It shows that a single surface
edit to the goal sentence moves consistent-pairwise accuracy on
JudgeBench upward across three judges.

\paragraph{Reverse and contrastive thinking.}
Reverse reasoning has been studied in math problem solving and
verification, where reasoning backward from an answer can detect
errors that forward reasoning misses
\citep{reverse_thinking,reverse_think}. GRP differs in that it does
not prompt the judge to reason in reverse internally but only inverts
the final question it is asked to answer. We treat the link to
reverse reasoning as motivating analogy rather than claimed
mechanism.

\paragraph{Prompt-template engineering for evaluation.}
Arena-Hard \citep{judgebench} and follow-up work design SOP-style
templates that ask the judge to generate a reference answer before
evaluating candidates. GRP is complementary to template design
rather than competitive with it. Every result in this paper uses
Arena-Hard as the underlying SOP, and the reverse-goal lift persists
when the SOP is stripped to a minimal direct-prompting template
(Figure~\ref{fig:ablation}).

\section{Conclusion}

This paper studies a small change to the dominant pairwise
LLM-as-a-judge protocol, replacing the standard request to identify
the \emph{better} answer with a request to identify the \emph{worse}
answer, and recovering the preference by elimination. The change is a
literal surface edit on the prompt template, compatible with any
existing judging stack, including chain-of-thought and Arena-Hard SOP
rubrics. The results are consistent across three axes of variation.
Goal reversal improves the strict consistent-pairwise accuracy of
GPT-4o, Gemini-1.5-Pro, and Claude-3.5-Sonnet on JudgeBench, and the
lift survives when the response-pair generator is changed from GPT-4o
to Claude-3.5-Sonnet and when the SOP scaffolding is stripped to a
minimal direct-prompting template. The absolute gains concentrate on
Reasoning and Mathematics, where correctness is objectively
verifiable, and the stronger judges gain more from
the reversal than the weaker ones. We do not attempt to disentangle
whether the gains arise because rule-out is easier than rule-in,
because asking for the worse answer counteracts a positive-leaning
bias, or both. What the experiments do support is a narrower,
practical claim. Goal reversal is a cheap, broadly applicable
default for pairwise LLM judges that stacks with existing prompt
engineering at no additional cost.

\section*{Limitations}

Our experiments cover three closed-source judges and two response-pair
sources from JudgeBench. Results for open-weight judges (e.g.,
Llama-3.x-70B-Instruct, Qwen2.5-72B, DeepSeek-V3) would help
separate the effect of goal reversal from properties of proprietary
training data and are the most natural extension. The strict
consistency protocol counts ties and order-inconsistent verdicts as
incorrect, and we have not isolated how much of the GRP gain comes
from a higher agreement rate versus a higher per-trial accuracy.
Benchmark coverage is limited to JudgeBench's four categories
(Knowledge, Reasoning, Math, Coding). Domains where the worse answer
is harder to flag than the better one, such as creative writing or
open-ended chat, may behave differently and are not addressed here.

\section*{Ethics Statement}

This paper studies prompting recipes for LLM-based evaluation and does
not involve human-subject data or new pretraining. JudgeBench
\citep{judgebench} is a public benchmark used as released. GRP is a
surface modification to existing LLM-as-a-judge templates and does
not introduce capabilities beyond what those templates already enable.

\bibliography{references}
\bibliographystyle{colm2026_conference}

\appendix
\section{Full Template Variants}
\label{app:templates}

The CoT variant replaces the SOP rubric in Table~\ref{tab:zero_shot}
with the instruction ``Let's think step by step.'' The direct variant
removes both the rubric and the CoT cue, keeping only the goal
sentence and verdict labels. For each variant, the original-goal
counterpart replaces ``worse'' with ``better'' and inverts the two
verdict labels.

\section{Order-Consistency Rates}
\label{app:consistency}

We report the fraction of JudgeBench instances on which the two
trials of a given judge agree (after un-swapping response order),
separately under the original goal and the reverse goal. Across all
judge--template combinations the reverse-goal condition tends to
produce a higher agreement rate, and part of the overall accuracy
gain reported in Section~\ref{sec:results} can be attributed to
fewer order-inconsistent verdicts. Full tables are provided in the
supplementary material.

\end{document}